# Paradigm Survey of Biology-inspired Spiking Neural Networks

Tianyu Zheng[1], Liyuan Han[1], Tielin Zhang[1,2],*

[1] Center for Excellence in Brain Science and Intelligence Technology, Chinese Academy of Sciences

[2] Institute of Automation, Chinese Academy of Sciences

*Corresponding author (zhangtielin@ion.ac.cn)

**Abstract**: Spiking neural networks (SNNs) are becoming increasingly popular in both brain simulation and brain-inspired computation due to their biological plausibility and computational efficiency. This paper reviews the historical development of SNNs and highlights how these two fields are progressively and interactively converging. With the successful deployment of dynamic vision and audio sensors, SNN algorithms have proven crucial in subsequent information processing for various SNN-friendly applications. These include continuous visual signal tracking, automatic speech recognition, and reinforcement learning for continuous control. In these contexts, key features of SNNs, such as spike encoding, neuronal heterogeneity, function-based neural circuits, and learning-enhanced neural plasticity, are prominently demonstrated. In addition to these machine-learning applications, there is another biological paradigm: invasive brain-computer interfaces (BCIs). BCIs involve decoding natural spike trains directly from the brain, without the need for additional spike encoding as required in traditional machine-learning paradigms. Since natural spike information from the brain closely mirrors the external world, BCIs align well with SNNs, allowing them to leverage their advantages in energy efficiency, robustness, and flexibility. This synergy between BCIs and SNNs not only enhances our understanding of the brain but also accelerates advancements in brain-inspired intelligence technology. This bidirectional interaction fosters significant progress in both brain science and the development of brain-inspired algorithms.

**Key Words**: Spiking neural network, Brain-computer interface, Experimental paradigm.

## 1 Introduction

As a brain-inspired neural network algorithm, the spiking neural network (SNN) [1] has garnered increasing attention from researchers in neuroscience and computer science since its introduction in 1997. SNNs offer both biological plausibility and computational efficiency, making them valuable in simulating biological brains and advancing brain-inspired computation. They bridge the gap between biological and artificial intelligence, enhancing our understanding of biological cognitive processes. After over 20 years of development, SNNs are widely used in computational neuroscience to simulate various biological scenarios. Simultaneously, they have demonstrated significant advantages in artificial intelligence applications, particularly in areas requiring high accuracy, energy efficiency, robustness, and flexibility. The future direction of SNNs, including the principles to uphold, compromises to make, and breakthroughs to pursue in application scenarios and model characteristics, represents key issues within the field of SNN research.

## 2 Evolutionary histories

SNNs, first introduced in 1997 as a novel type of ANN (often referred to as the third generation of ANNs[1]), emphasize spike encoding and neural dynamics as key mechanisms for information processing. Since their inception, SNNs have been compared to traditional ANNs across various dimensions, such as underlying mechanisms, learning methods, and evaluation metrics like accuracy and generalization. This comparison raises a fundamental question: after extensive genetic evolution, why did biological organisms favor SNNs over ANNs for intelligence? Beyond the obvious advantage of energy efficiency, are there intrinsic differences in neuron encoding, network structure, learning methods, or survival strategies that underpin this preference? Addressing these questions could significantly deepen our understanding of natural intelligence. Over the past two decades, SNNs have evolved significantly, primarily in two major application domains. The first focuses on biocomputational simulations grounded in computational neuroscience, aiming to elucidate multi-level perceptual and cognitive mechanisms in biological brains. The second domain centers on the innovative applications of ANNs within the framework of neuromorphic computing. Detailed explorations of these two directions are provided below.

## 2.1 Biocomputational simulations

Many biological mechanisms have been integrated into SNN algorithms to explain and validate the intelligent perception, working memory, and decision-making abilities of the biological brain (Fig.1). The Hodgkin-Huxley (H-H) model, introduced in 1952, followed the simpler leaky integrate-and-fire (LIF) model from 1907, which remains commonly used in conventional SNNs due to its straightforward mathematical formulation. By the 1990s, with the advent of neural morphology computation, SNNs emerged as a new ANN model based on asynchronous spike coding, gaining significant attention in fields like computer software and microelectronics. Neuromorphic hardware, which leverages these principles, holds the potential to surpass the traditional von Neumann computing architecture by offering advantages in integrated storage and computation, asynchronous parallel processing, and extremely low energy consumption. However, replicating the multi-scale computing capabilities of biology in artificial intelligence systems has proven more challenging than anticipated. The European Brain Project, launched in 2014 to simulate biological brain computation, has had to adjust its strategy due to increasing implementation difficulties. Even seemingly simple biological neurons and network structures have revealed complex nonlinear dynamics under advanced optical and electronic microscopy. In vivo recording techniques like two-photon imaging and patch clamping have uncovered long-term, multi-type neural plasticity mechanisms within networks. These include neuronal plasticity (such as dynamic discharge thresholds), synaptic plasticity (including spike-timing and short-term synaptic plasticity), and forms of meta-plasticity influenced by neurotransmitters like dopamine and acetylcholine.

After SNNs were introduced in 1997, they enabled the integration of various biological mechanisms within a single framework, simplifying the study of cognitive abilities and energy consumption in biological systems. SNNs, known as the third generation of ANNs, were distinguished from the first generation perceptron (a two-layer linear model) and the second generation multi-layer perceptron (a multi-layer nonlinear model). Numerous biological mechanisms and phenomena related to SNNs have since been uncovered through experimentation.

For example, in 2003, the stochastic information transmission mechanism of biological synapses was mathematically modeled and applied to synaptic learning in SNNs[2]. In 2006, the dynamic perturbation process of conductance was accurately simulated to enhance SNN learning[3]. In 2013, a specialized SNN network was first proposed to simulate insect decision control[4]. Between 2014 and 2017, various SNN models inspired by different brain regions were introduced, such as hippocampus-like multiscale ring SNN networks for robust memory formation and retrieval[5-7], and 6-layer cortical feature-constrained SNN networks for motion recognition[8].

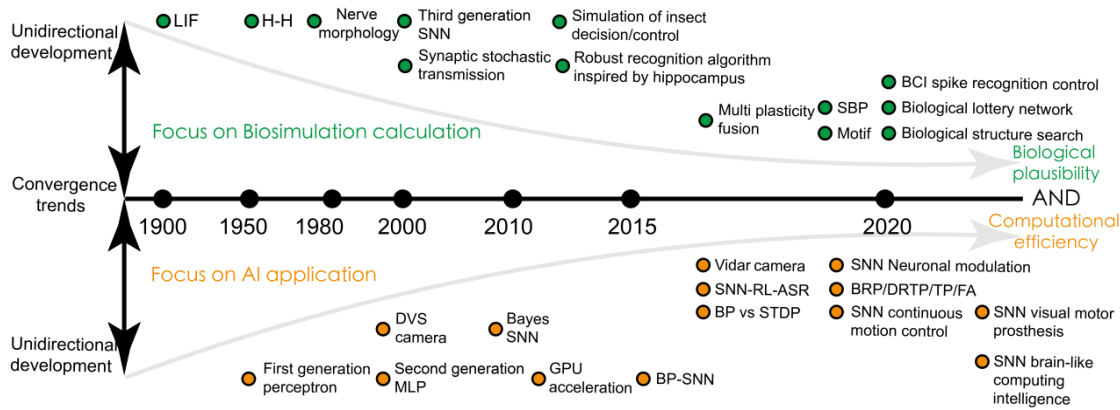

Fig. 1 Timeline of SNN historical development

After 2017, various biologically plausible SNN learning algorithms were developed, including the biologically inspired FORCE learning algorithm[9], multi-biological plasticity rule fusion algorithms[10-12], and the self-organizing backpropagation (SBP) learning algorithm[13]. New SNN models also emerged, inspired by biological structures, such as motif-based structure prediction[14], multisensory fusion algorithms based on motif loop structures[15], SNN architecture search via neural network structure search[16], and biological network evaluation using lottery theory[17]. These advances in biological computing offer valuable insights for constructing biologically plausible ANNs, marking a significant step towards next-generation AI algorithms. As more biological features are integrated into SNNs, they are gradually evolving from pure biological simulations to more AI-like applications, such as tasks like handwritten digit recognition with the MNIST dataset[18], digit speech recognition with the TIDigits dataset[17,19], speaker recognition using the TIMIT speech dataset[20], action recognition with the DvsGesture dataset[13], and reinforcement learning[20,21]. The convergence of biological simulation and AI applications is becoming increasingly evident, and fundamental questions about SNN applicability are coming to the forefront—such as which AI tasks can best leverage SNNs, and which are better suited for traditional ANNs.

## 2.2 Neuromorphic computing

The development of both SNNs and ANNs has been heavily influenced by evolving understandings of biological intelligence. Much of the existing SNN research treats SNNs as a spike-based variant of ANNs. Beyond spike encoding, other elements like network structure, learning methods, and optimizers often follow the same technical path as ANNs. The most extreme approach involves directly converting a well-trained ANN into an equivalent SNN[22]. This perspective aims to leverage the benefits of established ANN research, such as backpropagation and convolutional kernels, while enhancing energy efficiency and minimizing accuracy loss in SNNs. Guided by this approach, many SNN algorithms have been developed and applied in various specific scenarios[23,24].

Since the first dynamic vision sensor (DVS) was introduced in 2005[25], SNNs have been extensively studied as adaptive model algorithms. The DVS camera encodes changes in input into spike sequences, significantly boosting frame rate and reducing response time. Efforts to integrate Bayesian theory with SNN learning followed, enhancing their adaptability[26]. In 2012, SNN algorithms were first accelerated using GPUs[27]. By 2016, the use of backpropagation to optimize SNNs peaked, leading to the development and widespread application of deeper SNN structures in areas traditionally explored by ANNs[23,24].

During this period, the biological plausibility of backpropagation (BP) was revisited and widely discussed[28]. New biologically plausible BP variants were proposed, including Feedback Alignment (FA) [29], Biologically-plausible Reward Propagation (BRP)[18], Target Propagation (TP)[30], and Direct Random Target Propagation (DRTP) [31]. Even the classic Spike Timing-Dependent Plasticity (STDP) was shown to correlate with BP under energy function constraints[32], supporting the idea that BP might be biologically plausible. However, in biology, backpropagation operates in a self-organizing manner and is limited to adjacent layers, not extending beyond two layers[33,34]. As a result, a more realistic SBP algorithm was developed and validated on typical SNN and ANN networks[13]. In addition, non-gradient-based optimization methods, such as fidelity allocation methods inspired by dopamine reward learning[19] and various STDP variants[35], also played a crucial role in SNN learning. The continuous advancement of fundamental tools, including standardized Python toolkits[36], has significantly accelerated progress in the field, enabling the deployment of large-scale, deep SNNs in both software and hardware[37,38].

The SNN models mentioned above, designed for computational efficiency, have excelled not only in traditional tasks like image classification but also in processing DVS spike events, continuous dynamic speech recognition[39], and continuous action reinforcement learning[21,40]. As the accuracy of SNN models improves, researchers are increasingly focused on the biological characteristics of SNNs and their potential application advantages. The research trajectory of SNNs, initially driven by artificial intelligence applications, is now increasingly integrating biological plausibility with computational efficiency.

## 2.3 Towards BCI-based SNN

As discussed earlier, SNN development is increasingly trending towards cross-disciplinary fusion. Two main directions achieve both biological plausibility and computational efficiency: brain-inspired computing intelligence and SNN-based BCI decoding and regulation, the latter being the focus of this paper. Brain-inspired computing has advanced rapidly in typical computational tasks by integrating the strengths of ANNs and SNNs and incorporating new materials like memristors. For example, dynamic visual (DVS) and audio sensors (DAS) have achieved a balanced trade-off between computational accuracy and energy efficiency. In contrast, in BCI applications, the spike trains collected from invasive, high-throughput, and noisy electrode arrays are naturally suited for processing by SNNs, capitalizing on their speed, robustness, and flexibility.

SNNs, inspired by biological phenomena and mechanisms, offer significant computational advantages, particularly in analyzing biological signals in BCI. On one hand, brain-like SNN models align with multi-scale biological computing principles, providing clear interpretability for processing biological signals. On the other hand, SNNs have been used in computational neuroscience for simulations of brain functions, such as hippocampal learning and memory or visual perception. These models exhibit high biological plausibility, incorporating features like plasticity mechanisms, biological topologies, and diverse neuron types, making them suitable for future biological tissue simulations.

Moreover, in BCI signal regulation, SNNs hold distinct advantages. Their precise temporal characteristics make them ideal for continuous control tasks, such as reinforcement learning in Mujoco control[21]. For conditions like epilepsy, SNNs could potentially optimize brain activity control by calculating precise parameters for reverse electrical stimulation. In fields focusing on visual and motor prostheses, SNNs could directly interface with brain regions like the visual cortex (V1) or motor cortex (M1), enabling seamless integration with devices like cameras and robotic arms.

As biological plausibility and computational efficiency continue to converge, brain-like computing intelligence and BCI hardware and software are likely to integrate further, allowing BCI to fully leverage advancements in brain-inspired computing.

## 3 Key features of SNNs

The key distinction between SNNs and ANNs lies in the operator variations introduced by spike nodes and the differences in dynamic learning methods[41]. As shown in Fig. 2(a), ANNs typically use gradient-based learning methods like backpropagation (BP), where optimal synaptic adjustments are mathematically determined as errors propagate backward through the network. Recognizing the gap between BP and biological computation, several biologically plausible gradient methods have been proposed. These include Feedback Alignment (FA)[29], which allows error gradients to propagate without requiring exact weight symmetry; Target Propagation (TP)[28], which enables direct modulation of gradients across neuron states; and Equilibrium Propagation (EP)[11,12], which describes how local neuron states reach equilibrium under energy constraints. The evolution from BP to FA, TP, and EP reflects a trend towards increasingly biologically plausible gradient modifications.

Unlike the gradient-based learning methods used in ANNs, biological networks rely on neuroplasticity-driven learning. As shown in Fig. 2(b), by integrating various synaptic plasticity mechanisms across multiple scales, biological systems achieve efficient and reliable learning[10,42]. On a mesoscopic scale, neurotransmitters like dopamine globally influence synaptic plasticity, guiding different brain regions and circuits to perform specialized functions. On a microscopic scale, mechanisms such as Spike-Timing-Dependent Plasticity (STDP) adjust synaptic weights based on neuron activity; Short-Term Plasticity (STP) regulates input-output balance within synapses; and Synaptic Backpropagation (SBP)[33,43] explains how synaptic weights in the current layer are influenced by weight changes from neighborhood layers, through processes like Long-Term Potentiation (LTP) or Long-Term Depression (LTD). Beyond synaptic plasticity, neuron plasticity involves mechanisms like Homeostatic Thresholds (Homeo-V) and Adaptive Thresholds[11,13].

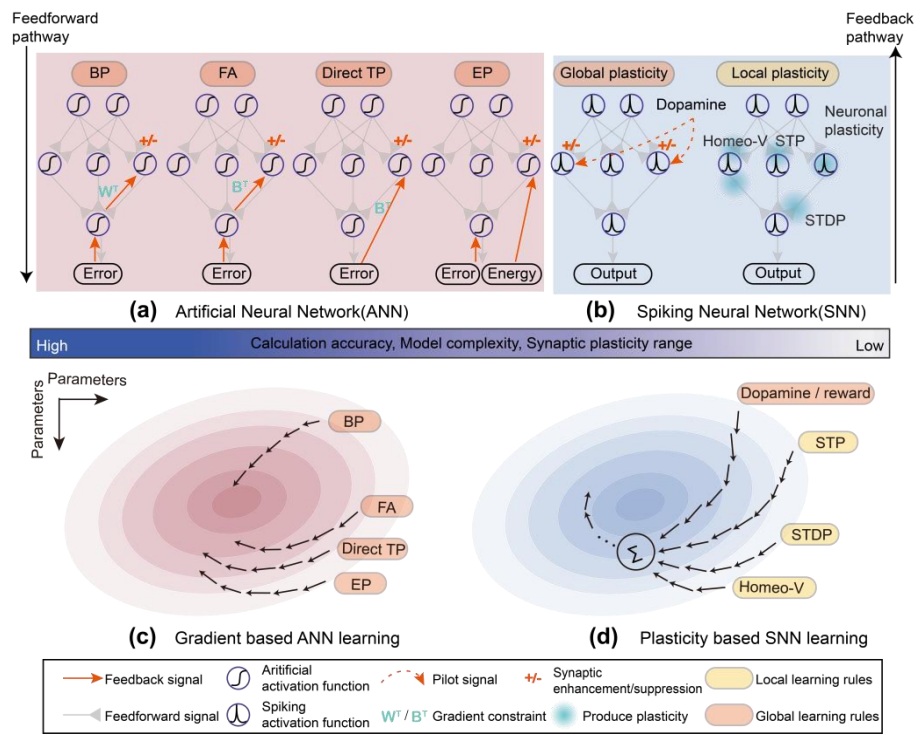

Fig. 2 Comparison of key characteristics between SNN and ANN

BP-based learning methods are the primary optimization strategy for ANNs, ensuring weight updates are mathematically aimed at minimizing global errors, as depicted in Fig. 2(c). In contrast, biological systems use a more approximate approach, integrating multiple optimization strategies that guide the network towards local Nash equilibrium states through self-organized local learning. Although these biological networks may be less effective in individual classification tasks compared to artificial BP methods, they still achieve convergence, as illustrated in Fig. 2(d). These networks offer advantages in algorithm complexity, energy efficiency, unsupervised learning, and adaptability[11,12].

The inclusion of spiking neurons, with their rich dynamic properties, introduces a critical time dimension to traditional operations like multiplication or convolution, thereby enhancing the diversity of temporal computations. As shown in Fig. 3(a), consider a standard convolution on a 2×2 input matrix (values ranging from 0 to 3) using a 2×2 kernel set to 1. Traditional methods would yield a result of 6 through matrix multiplication. However, in spiking convolution, the input is temporally encoded within a time window of 5. While the original 0 input remains unchanged (reused 5 times in the time domain), other values can be encoded in multiple temporal schemes. These different encoding schemes lead to variations in convolution outcomes and neuron integration processes, as shown in Fig. 3(b), potentially producing outputs of 5, 2, or other values. This variability underscores the importance of spike encoding in temporal processes, making SNN models particularly effective for tasks involving temporal information processing and continuous action control.

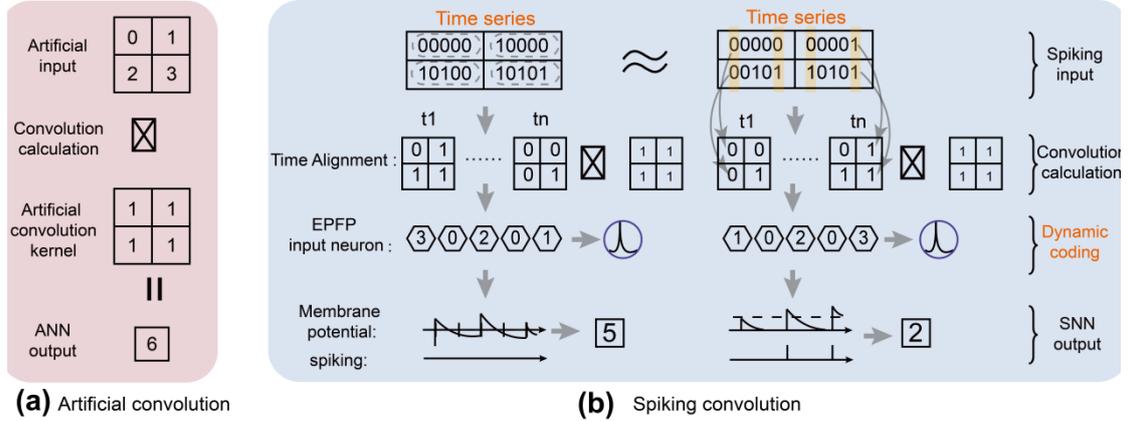

Fig. 3 Comparison of convolution computation between SNN and ANN

## 4 Three typical paradigms of SNNs

Most existing AI algorithms are designed to handle either static discrete signals or dynamic continuous signals, but the combination of both, i.e., dynamic discrete signals, is less commonly applied in traditional visual and auditory perception tasks. This is largely because these signals, characterized by events and asynchronous computation, complicate the learning process for software algorithms. With the emergence of SNN models, traditional visual and auditory devices have evolved into asynchronous DVS or DAS sensors, enabling intelligent perception in specific challenging environments, such as high-speed, low-light, or low-energy scenarios.

For example, DVS cameras offer extremely high frame rates compared to traditional optical cameras. Achieving similar frame rates with conventional cameras would require substantial investments in hardware, such as upgrading exposure systems or implementing multi-channel parallel processing. Asynchronous event signals, with their low latency and high encoding rates, are naturally well-suited for brain-like SNN algorithms. This section will explore the basic features of SNN models and their applications in visual information processing, auditory signal processing, and reinforcement learning for motion control.

### 4.1 Dynamic visual camera

DVS cameras capture dynamic scenes by focusing on changes in signals rather than continuous, invariant video streams. Using fixed or dynamic thresholding, they detect spike trains to encode these changes. Unlike traditional frame-based visual sensors, DVS outputs exhibit extremely low latency and a high dynamic range, complementing conventional cameras by enhancing visual quality and perceptual performance[44-47].

The low latency of DVS event spikes enables continuous monitoring of dynamic scenes, effectively capturing texture changes and motion information, which traditional cameras often miss due to issues like motion blur from long exposure times. By integrating dynamic spike data with traditional visual inputs, overall camera performance is significantly improved. For instance, a dual-path detection system mimicking the biological what and where pathways uses a beamsplitter to split visual information into two streams, optimizing the strengths of both ANN and SNN in processing natural video and spike data, as shown in Fig. 4.

In biological systems, visual information is processed through the retina, lateral geniculate nucleus, and primary visual cortex, eventually diverging into ventral (what) and dorsal (where) pathways. The dorsal pathway, which handles motion direction and spatial orientation, connects to the motion cortex through areas like MT, enabling rapid motion control or avoidance. Due to its focus on speed and temporal sensitivity, this pathway is particularly well-suited for SNN processing. SNNs leverage their precise timing and spike encoding capabilities to detect objects quickly across wide ranges and at high speeds.

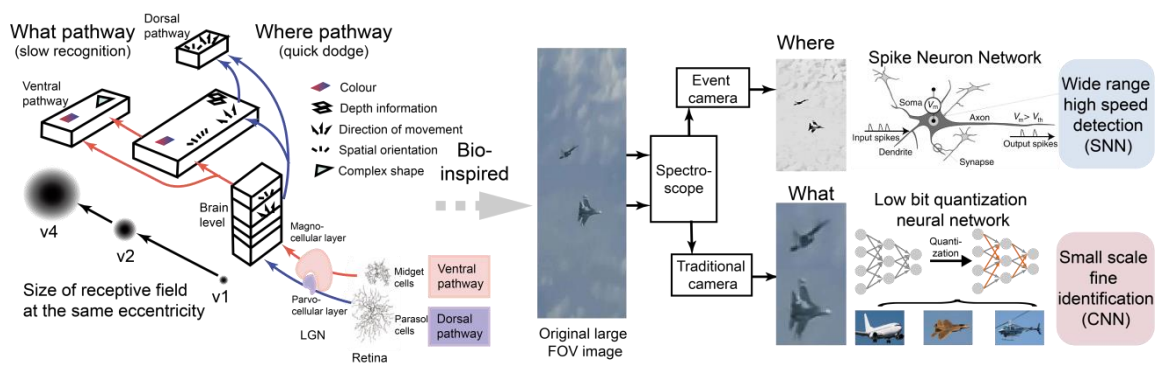

Fig. 4 SNN-based DVS signal processing[48]

The ventral pathway in vision focuses on analyzing and recognizing features like color, complex shapes, and textures, earning it the name what pathway. This pathway connects to advanced brain areas such as the prefrontal cortex and hippocampus via the inferotemporal (IT) cortex, enabling higher-level reasoning and memory analysis. ANN models are well-suited for processing this type of information, excelling at accurately identifying fine details through deep convolutional structures and techniques like low-bit quantization, compression, and distillation.

Each type of camera and corresponding algorithm has its strengths, particularly in processing speed and accuracy. A practical approach involves using SNNs with event cameras for rapid target detection, followed by ANNs with traditional cameras for detailed recognition. This mirrors the biological strategy of using the where pathway for quick motion localization and the what pathway for precise identification at a more deliberate pace. While numerous studies leverage the complementary strengths of both SNNs and ANNs, they are not elaborated here.

## 4.2 Auditory signal processing

A typical approach to processing spike-based auditory information involves encoding raw audio data into temporal-spatial spectral features, like Mel spectrograms, followed by key feature extraction and spike conversion[49,50]. These features are then used in Automatic Speech Recognition (ASR) modules. However, generating temporal-spatial spectrograms using digital filters or Fourier transforms is computationally intensive, posing challenges in resource-limited environments. Low-power Dynamic Auditory Sensors (DAS) can directly emulate the cochlea[51], performing analog band-pass filtering of sound information on-chip and encoding the spectral intensity into a continuous spike train. Despite this, the feature extraction capabilities of DAS are limited, necessitating further improvements in speech recognition performance.

In contrast to the frontend-backend model, the end-to-end model, which utilizes one-dimensional artificial or spike convolutions, can effectively learn and extract temporal-spatial features. By harnessing the dynamic temporal processing strengths of SNNs, this model shows great promise in auditory signal processing, offering a balance of low power consumption and high performance.

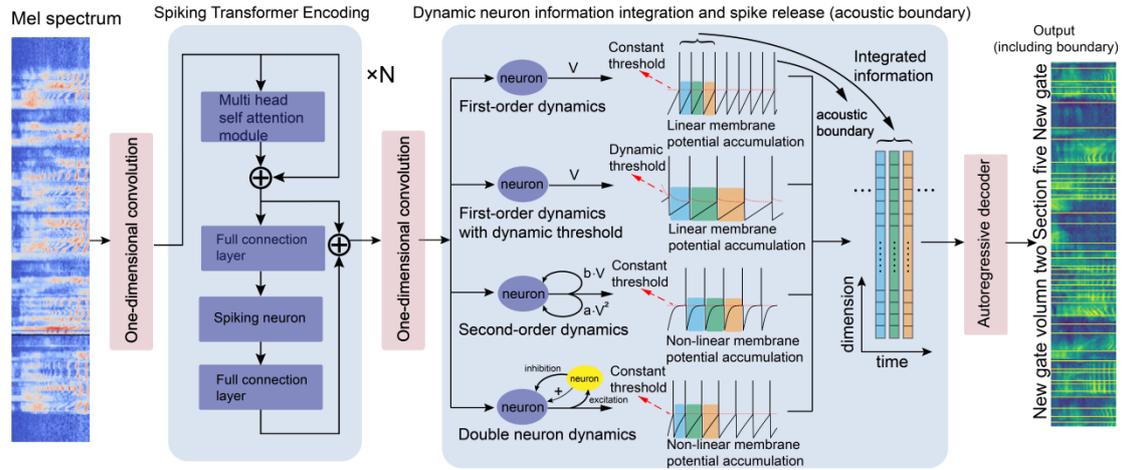

Fig. 5 SNN-based auditory information processing[52]

Fig. 5 illustrates a speech recognition approach that combines spike-based Transformers with dynamic neurons[52]. This method retains core components of traditional Transformers, such as multi-head attention modules and fully connected layers, but replaces the artificial ReLU activation with a spiking neuron layer for nonlinear processing. The spiking neurons introduce unique dynamic properties, such as first-order and higher-order dynamics, which help identify acoustic boundaries in the auditory information processing pipeline. This precise boundary extraction improves ASR accuracy by enhancing information segmentation in end-to-end network processing. Consequently, integrating spike-based acoustic boundaries with traditional speech features leads to higher phoneme and speech recognition accuracy, with reduced word error rates and lower energy consumption compared to ASR models using non-spike boundaries.

## 4.3 Reinforcement learning of continuous control

SNNs excel in tasks requiring continuous action control due to their sensitivity to temporal information[21,54], making them ideal for reinforcement learning environments like Mujoco. Mujoco features single-agent robots or unmanned vehicles that require continuous action outputs, aligning well with SNN's capabilities. In contrast, Atari tasks are less suited for SNNs. Atari games involve discrete actions (i.e., turning left, moving forward, or jumping), where temporal dynamics are minimal. These tasks often require additional sample buffers to handle random sampling and focus more on input-output mapping and fitting. Thus, the Atari environment is better suited for ANN models rather than SNNs.

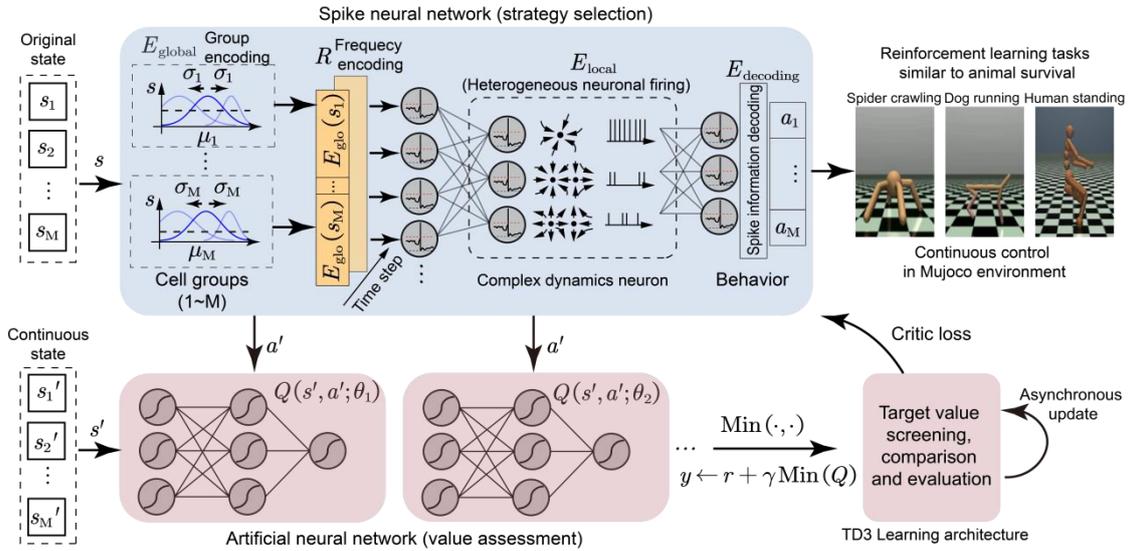

Fig. 6 SNN-based reinforcement learning in continuous control [21]

In Mujoco tasks, the input state vectors represent continuous control values for actions like joint movements and orientations. Similarly, the output action vectors correspond to control values for various joints or state quantities. These vectors are closely tied to temporal information, as agents learn to coordinate body parts for continuous control tasks. Examples include a quadruped learning to crawl faster (Ant-v3), a four-legged dog learning to run (HalfCheetah-v3), and a humanoid robot learning to stand (Humanoid-v3)[21]. More complex tasks, such as DeepMind's Humanoid Football task, involve multi-agent cooperative control for behaviors like kicking[55].

In Fig. 6, SNNs are used as spike policy networks for continuous action control in Mujoco. SNNs process raw state information and output continuous actions, leveraging their strength in handling temporal data. To address potential accuracy issues with event-based computation, neuron clusters are encoded, such as using multi-neuron group state elevation to convert raw states into spike trains. This encoding enhances the SNN's processing capabilities. Additionally, during training, an ANN may assist as a value evaluation network within traditional frameworks like TD3 or PPO to support the SNN policy network. In application, only the SNN policy network is used for continuous control. Mujoco's continuous action tasks mirror survival tasks in nature, making SNNs well-suited for these scenarios. This approach achieves a biologically plausible closed-loop from experimental design to model implementation.

## 5 BCI as a new SNN paradigm

Traditional tasks like dynamic visual processing, auditory analysis, and reinforcement learning are generally better suited to SNNs than ANNs due to their handling of temporal information. However, these methods are not without flaws. For instance, continuous data such as images and audio spectra need to be pre-encoded into spike trains before being processed by SNNs, leading to inherent information loss.

Although devices like DVS and DAS shift this encoding from software to hardware, they do not fully resolve the issue of temporal information loss.

A real-world application that naturally features spike trains, eliminating the need for additional temporal encoding, is the human brain. Neural spike trains in the brain directly reflect the external physical world, mapping complex information like videos, audio, and continuous controls to corresponding spikes, as shown in Fig. 7. With advancements in Brain-Computer Interfaces (BCI), encoding and decoding of brain data are transitioning from non-invasive methods (e.g., EEG, fMRI) to invasive ones (e.g., NeuroPixels), and from low-throughput to high-throughput synchronous recording.

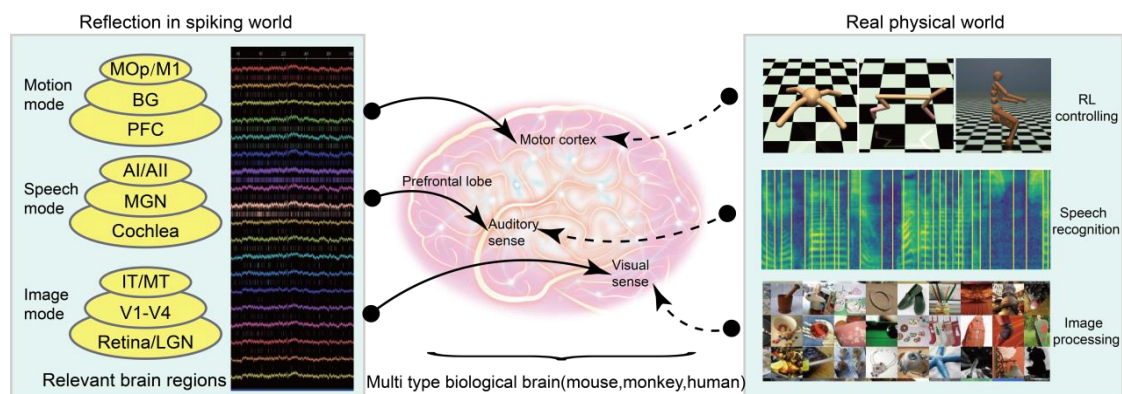

Fig. 7 Natural spike trains in invasive BCI paradigm fit for SNNs

The brain's complex dynamics produce numerous spike trains to encode multimodal information from the physical world. These spike trains, generated by various brain regions, capture different levels of abstraction across sensory modalities like vision, hearing, touch, and movement. To decode these patterns, models need capabilities akin to those used for processing DVS and DAS event data. Additionally, the prevalence of spike noise adds to the complexity of information processing. This chapter introduces a new experimental paradigm using BCI with SNNs, highlighting the advantages of SNN models in handling such complex data.

## 5.1 Mouse-BCI paradigms

The spike working memory and cognitive decision-making task based on the mouse delayed matching-to-sample (ODPA) paradigm[56] is an ideal example of using SNNs, as shown in Fig. 8(a). In this classic task, mice are exposed to paired or non-paired odor inputs during a delay period. The length of this delay, ranging from milliseconds to several minutes, significantly affects the mice's decision-making. The task engages various brain regions, including the sensory cortex, hippocampus, and medial prefrontal cortex, reflecting pathways for sensory perception and decision-making.

The SNN model's strength lies in its ability to handle the temporal dynamics of these spike trains, which are crucial for maintaining working memory across the delay period. The experimental setup alternates delay intervals of 3 and 6 seconds, with sample and test odor times of 1 second each, and a response time of 1 second. Correct responses are indicated by licking water for paired trials (Hit) or refraining from licking for non-paired trials (Correct Rejection). Incorrect responses include missing water during paired trials (Miss) or licking water during non-paired trials (False Alarm).

This ODPA task requires mice to accurately pair sample and test odors, with decision-making and movement planning occurring only after the test odor appears. The task's design ensures clear trial structures and minimal licking behavior during the delay period. By integrating spatial (odor representation) and temporal (working memory maintenance) information into spike trains, the ODPA task highlights the SNN model's ability to manage complex spatiotemporal data. The 3s and 6s delay periods are crucial for decoding time-coded signals from extensive spike sequences, emphasizing the need for SNN-based decoding algorithms to classify and reconstruct these data effectively.

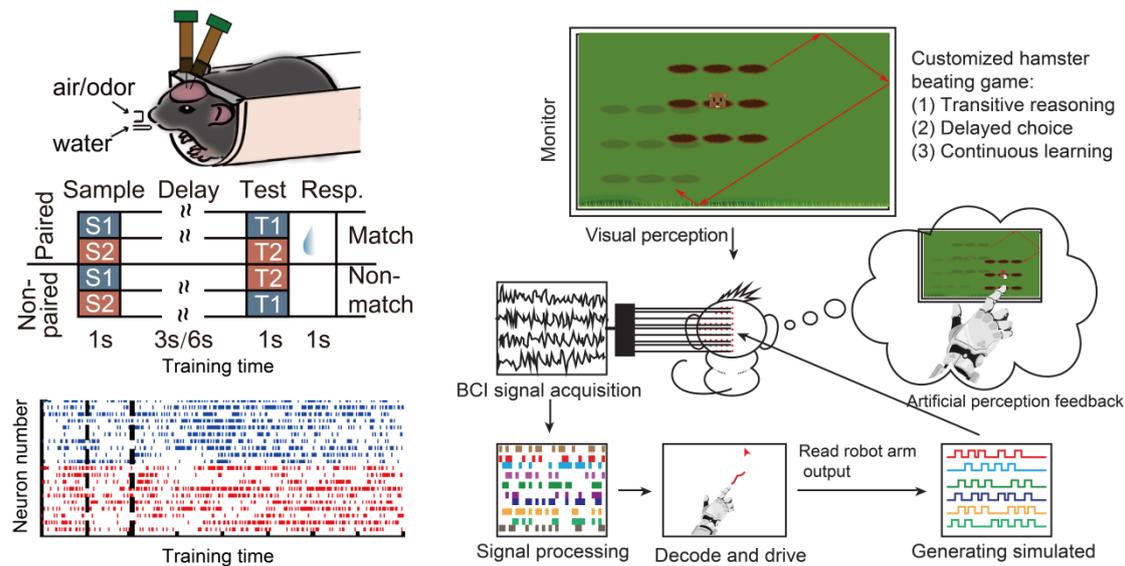

Fig. 8 Two typical BCI paradigms in mouse and monkey brains

## 5.2 Monkey-BCI paradigms

The spike-based continuous motion control task, exemplified by the macaque's free grasping task, is a key SNN paradigm. This task can vary, including fixed point tasks like the center-out task or flexible reach tasks. In these experiments, macaques use visual cues to guide their hand movements, receiving rewards for correct actions and penalties for errors. The task involves synchronous electrode recording from brain areas such as the motor cortex and premotor cortex, capturing complex spike trains during control planning and execution phases. The macaque's hand movement trajectory is also recorded to benchmark peripheral control.

In the realm of BCI, motion control paradigms are valuable for testing SNN applications. Traditional paradigms often focus on simple, rigid movements for decoding kinematic parameters and controlling peripherals like robotic arms. However, free and flexible manual motion control has been less explored due to limitations in linear neural decoding algorithms for complex motion states. The macaque touchscreen task, like the whack-a-mole game, offers a model for such exploration.

In this task, the macaque touches a start signal on the screen, then responds to randomly appearing moles within a set time. The hand movements are not constrained, and mole positions may change continuously. This setup enhances motion control flexibility. By leveraging brain-forward prediction and closed-loop feedback with electrical stimulation, SNN decoders can excel in controlling such free-motion tasks, demonstrating their potential for advanced motion control applications.

### 5.3 SNN-based BCI decoding

By collecting high-throughput EEG data, we can record spike discharge processes in areas related to perception, working memory, and decision-making. This enables visualization of spike train patterns as transient memory trajectories. After standardizing, aligning, and filtering the data, we create a dataset suitable for SNN applications involving working memory, recognition, and control.

The process of using SNNs for neural signal processing begins with data collection through invasive BCI techniques. For example, a BCI system with 5,000 channels records spike trains from various brain regions like the premotor cortex, hippocampus, prefrontal cortex, motor cortex, and thalamus[58]. This provides a comprehensive dataset from monkeys performing specific tasks. Raw neural signals are then preprocessed to enhance signal quality and reduce noise, typically using band-pass filtering to remove unwanted frequencies and direct current offsets. For instance, a band-pass filter with a 300-3000 Hz range can be used[59].

The next step is spike detection and sorting, which involves identifying and classifying spikes within the preprocessed signals. Accurate spike detection is crucial for converting continuous recordings into discrete spike trains. Common algorithms include amplitude thresholding, template matching, and wavelet transformation. Amplitude thresholding with adaptive thresholds is often chosen for its simplicity and real-time efficiency. SNNs process the detected spikes by mimicking biological neural processes, making them ideal for handling temporal dynamics. SNNs encode information using discrete spikes, offering high temporal resolution. The LIF model, a widely used neuron model in SNNs, simulates the integration of input spikes over time until a threshold triggers an output spike. This model effectively captures temporal features of neural data, and research shows that SNNs can simulate both microscopic spikes and mesoscopic population firing rates.

In summary, SNN processing involves data collection, preprocessing, spike detection, and implementation of neuron models. These steps improve our understanding and utilization of neural signals, advancing neuroscience and related fields.

## 6 Summary and Discussion

This paper starts with a historical overview of SNNs, examining their evolution and progress in biologically inspired computing and artificial intelligence. It notes a trend towards convergence between these domains and suggests that invasive, high-throughput BCI could be the next major development, integrating analog processing, recognition, and control of spike sequences.

The paper then highlights the unique capabilities of SNNs in processing temporal and continuous information, detailing their application in three key task areas: dynamic vision camera, auditory signal processing, and continuous control.

Finally, it explores spike train recognition and control within the BCI framework,

focusing on experimental paradigms such as working memory decision-making, motor control, and biological studies involving mice and monkeys. It compares complex brain information processing with classic external tasks.

The paper concludes that scenarios involving high-throughput complex spike trains, akin to or more complex than external tasks like dynamic audiovisual recognition and continuous motion control, could serve as ideal validation environments for future SNN applications. SNNs offer significant advantages in spike encoding, structural complexity, and biological plausibility. Additionally, neural morphological devices incorporating SNN cognitive functions hold promise as core components for BCIs, potentially enhancing visual or motor prostheses and opening exciting future possibilities.